# Shiva: A Framework for Graph based Ontology Matching

Iti Mathur, Nisheeth Joshi
Apaji Institute
Banasthali University
Rajasthan, India

Hemant Darbari, Ajai Kumar
Applied Artificial Intelligence Group
Center for Development of Advanced Computing
Pune, Maharashtra, India

## ABSTRACT
Since long, corporations are looking for knowledge sources which can provide structured description of data and can focus on meaning and shared understanding. Structures which can facilitate open world assumptions and can be flexible enough to incorporate and recognize more than one name for an entity. A source whose major purpose is to facilitate human communication and interoperability. Clearly, databases fail to provide these features and ontologies have emerged as an alternative choice, but corporations working on same domain tend to make different ontologies. The problem occurs when they want to share their data/knowledge. Thus we need tools to merge ontologies into one. This task is termed as ontology matching. This is an emerging area and still we have to go a long way in having an ideal matcher which can produce good results. In this paper we have shown a framework to matching ontologies using graphs.

## 1. INTRODUCTION
Since the dawn of Semantic Web. Ontology Matching (OM) is gaining popularity. As corporations have started using ontologies for storing their knowledge. This knowledge is the most valuable asset for any organization and timely access to this knowledge is the major focus to them. Unfortunately this is not as simple as it sounds because at times a knowledge engineer has to come across with a situation where more than one ontology is being used for the same knowledge. This is a nightmare which every knowledge engineer fears.

To address this issue, one has to either employ a human annotator to merge all the ontologies having same knowledge or they have to devise a mechanism to merge the ontologies automatically. This latter part is termed as ontology matching. Since the beginning of $21^{st}$ century this concept is being widely explored. Researchers are trying to develop new ways to merge ontologies which can produce results as good as humans. The problem of merging or matching ontologies is not as simple, as there are several issues that are to be considered while matching ontologies. Among them the most prominent issue is of heterogeneity where ontologies are available in different frameworks and we need to merge the knowledge incorporated in them. Most of the matchers developed today are unable to handle this problem. In our approach we have addressed this issue. As mostly the ontologies are available in OWL, RDF or XML formats. Our matcher can read any of these formats and can align their information and produce an aligned ontology.

The rest of the paper is as follows: Section 2 gives a brief description of the work done in the area of ontology matching. Section 3 describes our approach; it explains the experimental setup and our methodology. Section 4 describes the evaluation procedure incorporated to test the performance of the matcher and Section 5 concludes the work done.

## 2. LITERATURE SURVEY
In the past decade, as this area gained popularity, a lot of work was done to develop good matching systems. In this section we describe some of the best matchers developed till date.

Agreement Maker is a matcher developed at University of Illinois at Chicago by Cruz at el. [1]. This system has the best user interface developed so far. Moreover it has a flexible architecture and an integrated user interface which makes it different from other matchers. The core philosophy of the developers of this matcher is that of involving users into matching process. They believe that users can help make better alignments which are not possible in automatic alignments. Thus they prophesize the use of having semi-automatic matching systems. LogMap is another ontology matcher which is developed at University of Oxford by Ruiz and Grau [2]. They have used a logic based reasoning approach in their matcher. Their argument is that if we use logic based semantics incorporated in the ontologies then we may produce better alignments. This matcher is still under development and has already started a debate among the circles of the developers of ontology matchers.

AROMA [3] is a hybrid ontology matcher which can effectively match the concepts and properties from two ontologies. In order to do so they use association rule paradigm [4] and statistical interestingness measure. CIDER [5] tries to align ontologies using schema matching. It follows a two pronged approach, first it tries to extract similar concepts up to a certain depth and then applies different matching techniques onto the concepts and then finally produce aligned ontology. Lily [6] is another matching system which has re-emerged as one of the active ontology matchers. It can match generic and large scale ontologies. It can produce good results for normal size ontologies but it takes a lot of time to do so. The main reason behind this is that this matchers tries to extract semantic sub graphs and then tries to map them with other ontologies.

RiMOM [7] is one the top performing matchers that are tested in various evaluation campaigns across the globe. This matcher can not only match schema but also can match instances available in the ontologies. It uses multiple techniques to implement this feature and uses external resources like WorldNet to do semantic matching. TaxoMap [8] is another matcher which can produce matched ontologies of large scale. It does so by finding correspondence between the concepts of two ontologies. It also performs matches for subsumption relations and its inverse and proximity relations. YAM++ [9] is another matcher which can produce good results. This system uses multiple matching algorithms which are combined to produce matched ontology. This system provides flexibility as it allows the user to provide preferences. This system is self-configurable and extensible as if the user is not satisfied with the results then he can provide his own customized matching approach.

## 3. OUR APPROACH
### 3.1 Experimental Setup
To test the performance of ontology matchers, we required ontologies. So, we used some of the ontologies with OAEI (Ontology Alignment and Evaluation Initiative) 2013 evaluation task [10]. This task had some lightweight ontologies and one heavyweight ontology. We used fifteen ontologies from benchmark test set. These were light weight ontologies. We also used an ontology from anatomy track.

Since we could not find any more heavy weight ontologies, we developed some ontologies on our own. These were ontology on human anatomy [11] which had concepts relating to human physiological structure; we also developed two ontologies on health care services [12] and communicable diseases [13].

We have also used some of the best matchers from the OAEI 2013 task and compared our system with them. We used a graph based methodology for matching the ontologies. The objective was to check the feasibility of graph matching algorithms into ontology matching. Although some work has been done for using graphs in ontology matching. None of the previous work has checked the feasibility of graph based matchers with both heavyweight as well as lightweight ontologies.

### 3.2 Methodology
As ontologies have a hierarchical structure where concepts, attributes and instances can be arranged in a tree/graph like structure; using a graph matching algorithm here is far more intuitive mechanism. Thus, in our approach we have done the same. We have used bi-partite graph matching algorithm in our approach.

We have christened our system as Shiva. In our approach, we first take two ontologies. These can be in different formats. For example, the source ontology can be in OWL format while the target ontology can be in RDF format. Our system can recognize ontologies in OWL, RDF and XML formats. So, the source ontology $O_s$ and target ontology $O_t$ are read and are sent for preprocessing. In preprocessing task, first we separately parse the ontologies by collecting various concepts, sub-concepts, properties and instances. This information is stored in a file for manual debugging. Moreover, this extracted information is preprocessed and is arranged into a linked graph in memory. Thus each concept has a direct relationship with its properties, sub-concepts and instances. If we want we can generate an adjacency metric of this information or we can see it visually by creating vertices and arcs labeled as Isa, instanceof and hasproperty.

Once preprocessing is completed, the extracted information is sent to the matching system, were the user has the choice to selection from four different structural matching algorithms these are: Levensthein Edit Distance [14], Qgrams [15], Smith Waterman [16] and Jaccard's Coefficient [17] algorithms. All the algorithms search for similarities between concepts, sub-

concepts, properties and instances and are checked for three types of correspondences. These are:

1.  Equivalence correspondence: where a concept, sub-concept, property or instance in $O_S$ matches with its counterpart (at same level) in $O_t$.

2.  Isa correspondence: where a sub-concept of $O_S$ matches with a concept of $O_t$ and vice versa.

3.  General correspondence: where a property $O_S$ matches with a concept or sub-concept of $O_t$ and vice versa.

Thus all the mapping (mapping$(x, y)$) are generated using four tuples$(x, y, r, t)$. Where:

$x \in O_s$: x belongs to concepts, sub-concepts, properties and instances in source ontology.

$y \in O_t$: y belongs to concepts, sub-concepts, properties and instances in target ontology.

$r \in R$ : r is a correspondence relations in a set of correspondence relations R, in our case these are Equivalence, Isa and General.

t $\in$ T : t is the similarity metric used in alignment from a set of available metrics T, in our case these are Levensthein Distance, Jaccard Coefficient, Smith Waterman and Qgrams.

Using these mappings, we generated a score matrix in the following format:

$$S = \begin{bmatrix} M[O_{11}O_{21}] & M[O_{12}O_{21}] & M[O_{13}O_{21}] \dots M[O_{1m}O_{21}] \\ M[O_{11}O_{22}] & M[O_{12}O_{22}] & M[O_{13}O_{22}] \dots M[O_{1m}O_{22}] \\ M[O_{11}O_{23}] & M[O_{12}O_{23}] & M[O_{13}O_{23}] \dots M[O_{1m}O_{23}] \\ \vdots & \vdots & \vdots & \vdots \\ \vdots & \vdots & \vdots & \vdots \\ M[O_{11}O_{2n}] & M[O_{12}O_{2n}] & M[O_{13}O_{2n}] \dots M[O_{1m}O_{2n}] \end{bmatrix}_{m \times n}$$

Here, $M[O_{11}O_{21}]$ is the mapping between one of the elements (concepts, sub-concepts, properties, instances) of source ontology $O_S$ with one of the elements (concepts, sub-concepts, properties, instances) of target ontology $O_t$. This has the value which is produced by the similarity metric. For example, if we are using levensthein distance algorithm and we have two concepts as car and cars, then its score would be 1 and the similarity is calculated using the formula in equation 1.

$$sim(x, y) = \frac{\#edits\,(x, y)}{\max\,(len\,(x), len\,(y))} \qquad (1)$$

Here x and y are the two strings, in our case x is "car" and y is "cars". #matches(x,y) is the no. of edits required to make the two strings equal and len(x) is the length of string x, len(y) is the length of string y. the maximum of the two is selected to compute the final score. This is done for all the mappings which then generate the score matrix of all the matched elements of both the ontologies. This matrix can be seen as bipartite graph which has two disjoint sets of vertices (in our case mapping elements of $O_S$ and $O_t$) and edge weights (similarity values) are clearly mentioned.

Once the score matrix is generated, it is passed to our graph matching algorithm. We used Hungarian method [18] for matching our score matrix (bipartite graph). This gave us the best matching pairs in the matrix which are then used to generate the aligned ontology. Figure 1 shows the architecture of our system. A snapshot of aligned ontology is shown in figure 2.

## 4. EVALUATION

To evaluate the performance of our system we used 19 ontologies. Among them 15 were light weight ontologies and 4 were heavy weight ontologies. We used 2 popular ontology matchers (RiMOM and YAM++) with our four variants and compared their performance. We performed our evaluation on three categories. In first category we matched all the ontologies. In the second category we only matched the light weight ontologies and in the third category we only matched the heavy weight ontologies. We calculated precision, recall and f-measures using equations 2, 3 and 4 respectively.

$$Precision \ (P) = \frac{\#correct \ \_mappings}{\#total \ \_mappings \ \_system} \qquad (2)$$

$$Recall \ (R) = \frac{\#correct \ \_mappings}{\#total \ \_mappings \ \_human} \qquad (3)$$

$$F - Measure \ (F) = \frac{2 \times P \times R}{P+R} \qquad (4)$$

Here, the system generated matched ontology is compared with human generated manually matched ontology. The basic idea is to make the system produce an ontology which can emulate human matched ontology. Thus the matchers which produce better mappings is considered being the best. Precision is calculated using the correct mappings between human and system's ontology divided by the total mappings produced by the system. Recall is calculated using the correct mappings between human and system's ontology divided by the total mappings produced by the human. F-measure is the combination of the two.

Table 1 shows the values of precision, recall and fmeasure. While taking the average of all the ontologies, we found that RiMOM performed better than all other matchers while Shiva with Levensthein Distance Algorithm was second. In category 2, where only light weight ontologies were considered, we computed the averages of only these ontologies and found that again RiMOM performed better than other matchers with Shiva with Levensthein Distance Algorithm managing to get the second position. For category 3, we only took the averages of heavy weight ontologies and found that RiMOM again was the top matchers. This time YAM++ performed better than Shiva with Levensthein Distance Algorithm.

## 5. CONCLUSION

In this paper, we have shown the implementation of a graph based matcher with four different variants which use four different algorithms. We have used bipartite graph matching algorithm in creating aligned ontology. This approach produced good results as it could work at par with YAM++, one of the good ontology matchers while could not match with RiMOM. One of the reasons for this is that RiMOM matches ontologies at semantic level while Shiva only matches them at structural level.

As an enhancement to this work, we can add WorldNet and similar semantic resources to improve the performance of the matcher by combining structural as well as semantic matching techniques.

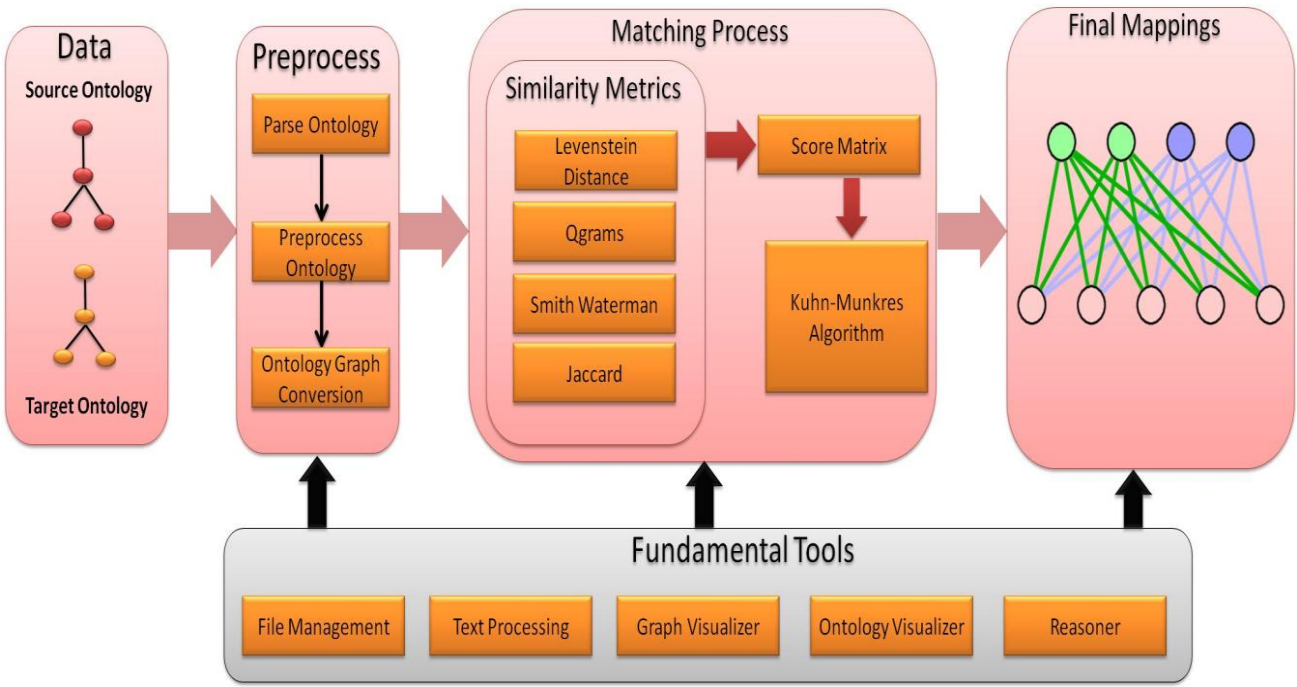

**Figure 1: Architecture of Shiva Ontology Matching System and Framework**

```
− <rdf:RDF xml:base="http://knowledgeweb.semanticweb.org/heterogeneity/alignment#">
  − <Alignment>
      <xml>yes</xml>
      <level>0</level>
      <type>11</type>
      <method>Automated generated</method>
   + <onto1></onto1>
   + <onto2></onto2>
      <uri1>http://www.w3.org/2001/XMLSchema/O1.rdf</uri1>
      <uri2>http://www.w3.org/2001/XMLSchema/O2.rdf</uri2>
    − <map>
      − <Cell>
          <entity1 rdf:resource="http://www.w3.org/2001/XMLSchema/O1.rdf#Academic"/>
          <entity2 rdf:resource="http://www.w3.org/2001/XMLSchema/O2.rdf#Academic"/>
          <measure rdf:datatype="http://www.w3.org/2001/XMLSchema#float">1.0</measure>
          <relation>=</relation>
        </Cell>
      </map>
    − <map>
      − <Cell>
          <entity1 rdf:resource="http://www.w3.org/2001/XMLSchema/O1.rdf#Book"/>
          <entity2 rdf:resource="http://www.w3.org/2001/XMLSchema/O2.rdf#Book"/>
          <measure rdf:datatype="http://www.w3.org/2001/XMLSchema#float">1.0</measure>
          <relation>=</relation>
        </Cell>
      </map>
```

**Figure 2: Snapshot of aligned ontology**

**Table 1: Comparison of Evaluation Results**

| Ontology | RiMOM | | | YAM++ | | | Shiva$_{Jaccard}$ | | | Shiva$_{LD}$ | | | Shiva$_{Dgram}$ | | | Shiva$_{tew}$ | | |
|---|---|---|---|---|---|---|---|---|---|---|---|---|---|---|---|---|---|---|
| | P | R | F | P | R | F | P | R | F | P | R | F | P | R | F | P | R | F |
| 101 | 1 | 0.98969 | 0.99481 | 0.75257 | 0.42941 | 0.54681 | 0.59036 | 0.50515 | 0.54444 | 0.975 | 0.80412 | 0.88135 | 0.13725 | 0.64948 | 0.22661 | 0.92957 | 0.68041 | 0.7857 |
| 103 | 0.96875 | 0.95876 | 0.96373 | 0.92783 | 0.48128 | 0.63380 | 0.55056 | 0.50515 | 0.52688 | 0.94666 | 0.73195 | 0.82558 | 0.11623 | 0.64948 | 0.19718 | 0.92957 | 0.68041 | 0.7857 |
| 104 | 0.96875 | 0.95876 | 0.96373 | 0.92783 | 0.48128 | 0.63380 | 0.55056 | 0.50515 | 0.52688 | 0.95876 | 0.95876 | 0.95876 | 0.11623 | 0.64948 | 0.19718 | 0.92957 | 0.68041 | 0.7857 |
| 201 | 0.90909 | 0.72164 | 0.80459 | 0.92783 | 0.48128 | 0.63380 | 1 | 0.05825 | 0.11009 | 1 | 0.98969 | 0.99481 | 0.06862 | 0.06730 | 0.06796 | 0.0845 | 0.06185 | 0.0714 |
| 201-2 | 0.86111 | 0.63917 | 0.73372 | 0.92783 | 0.48128 | 0.63380 | 0.61702 | 0.29896 | 0.40277 | 0.97894 | 0.95876 | 0.96875 | 0.11421 | 0.50515 | 0.18631 | 0.76056 | 0.5567 | 0.6425 |
| 201-4 | 0.92 | 0.71134 | 0.80232 | 0.92783 | 0.48128 | 0.63380 | 0.57142 | 0.16494 | 0.256 | 0.97894 | 0.95876 | 0.96875 | 0.11111 | 0.38144 | 0.17209 | 0.54929 | 0.40206 | 0.4642 |
| 201-6 | 1 | 0.82474 | 0.90395 | 0.92783 | 0.48128 | 0.63380 | 0.66666 | 0.08247 | 0.14678 | 0.97894 | 0.95876 | 0.96875 | 0.10038 | 0.26804 | 0.14606 | 0.39436 | 0.28866 | 0.3333 |
| 201-8 | 1 | 0.86597 | 0.92817 | 0.92783 | 0.48128 | 0.63380 | 1 | 0.04902 | 0.09345 | 0.93333 | 0.57732 | 0.71337 | 0.07471 | 0.13402 | 0.09594 | 0.23943 | 0.17525 | 0.2023 |
| 202 | 1 | 0.86597 | 0.92817 | 0.92783 | 0.48128 | 0.63380 | 1 | 0.06730 | 0.12612 | 1 | 0.91752 | 0.95698 | 0.07767 | 0.07619 | 0.07692 | 0.0845 | 0.06185 | 0.0714 |
| 202-2 | 1 | 0.84536 | 0.91620 | 0.92783 | 0.48128 | 0.63380 | 0.61702 | 0.29896 | 0.40277 | 0.98795 | 0.84536 | 0.91111 | 0.11421 | 0.50515 | 0.18631 | 0.76056 | 0.5567 | 0.6428 |
| 202-4 | 1 | 0.84536 | 0.91620 | 0.92783 | 0.48128 | 0.63380 | 0.57142 | 0.16494 | 0.256 | 0.98717 | 0.79381 | 0.88 | 0.11111 | 0.38144 | 0.17209 | 0.54929 | 0.40206 | 0.4642 |
| 202-6 | 1 | 0.87628 | 0.93406 | 0.92783 | 0.48128 | 0.63380 | 0.66666 | 0.08247 | 0.14678 | 0.97260 | 0.73195 | 0.83529 | 0.10038 | 0.26804 | 0.14606 | 0.39436 | 0.28866 | 0.3333 |
| 202-8 | 0.91304 | 0.64948 | 0.75903 | 1 | 0.5 | 0.66666 | 1 | 0.07619 | 0.14159 | 0.72340 | 0.70103 | 0.71204 | 0.07471 | 0.13402 | 0.09594 | 0.23943 | 0.17525 | 0.2023 |
| 203 | 1 | 0.77319 | 0.87209 | 0.92783 | 0.48128 | 0.63380 | 0.55056 | 0.50515 | 0.52688 | 0.72340 | 0.70103 | 0.71204 | 0.11623 | 0.64948 | 0.19718 | 0.92957 | 0.68041 | 0.785 |
| 204 | 1 | 0.77319 | 0.87209 | 0.92783 | 0.48128 | 0.63380 | 0.56097 | 0.47422 | 0.51396 | 0.72340 | 0.70103 | 0.71204 | 0.12403 | 0.65979 | 0.20880 | 0.88732 | 0.64948 | 0.75 |
| Anatomy | 0.97222 | 0.72164 | 0.82840 | 0.65591 | 0.39610 | 0.49392 | 0.58181 | 0.34408 | 0.43243 | 0.89743 | 0.37634 | 0.53030 | 0.11428 | 0.64516 | 0.19417 | 0.93846 | 0.65591 | 0.77215 |
| OntoAna | 0.97058 | 1 | 0.98507 | 0.92783 | 0.48128 | 0.63380 | 0.50549 | 0.47422 | 0.48936 | 0.952381 | 0.412371 | 0.57554 | 0.07588 | 0.63917 | 0.13566 | 0.91549 | 0.6701 | 0.77381 |
| HlthCare | 0.98795 | 0.84536 | 0.91111 | 0.2414 | 1.05197 | 2.21928 | 0.60869 | 0.48275 | 0.53846 | 1 | 0.27272 | 0.42857 | 0.14110 | 0.79310 | 0.23958 | 0.85185 | 0.7931 | 0.82142 |
| HCD | 1 | 0.7628 | 0.865497 | 0.78787 | 0.44067 | 0.56521 | 0.5333 | 0.72727 | 0.61538 | 1 | 0.2727 | 0.4285 | 0.05921 | 0.81818 | 0.11042 | 0.81818 | 0.81818 | 0.81818 |
| **Average Category1** | **0.9721** | **0.8225** | **0.8885** | 0.8645 | 0.5029 | 0.7034 | 0.6706 | 0.3087 | 0.3577 | **0.9325** | **0.7191** | **0.7875** | 0.1025 | 0.4670 | 0.1606 | 0.6413 | 0.4882 | 0.5529 |
| **Average Category2** | **0.9693** | **0.8199** | **0.8861** | 0.9209 | 0.47907 | 0.6301 | 0.7008 | 0.2558 | 0.3147 | **0.9245** | **0.8219** | **0.8666** | 0.1038 | 0.3985 | 0.1581 | 0.5774 | 0.4226 | 0.4879 |